\documentclass{midl} 

\usepackage{mwe} 
\usepackage{multirow}
\usepackage{enumitem}

\jmlryear{2025}
\jmlrworkshop{Full Paper -- MIDL 2025}
\jmlrvolume{-- nnn}
\editors{Accepted for publication at MIDL 2025}

\title[RaDialog]{RaDialog: Large Vision-Language Models for X-Ray Reporting and Dialog-Driven Assistance}

\midlauthor{\\
\Name{Chantal Pellegrini\nametag{$^{1,3}$}} \Email{chantal.pellegrini@tum.de}\\
\Name{Ege Özsoy\nametag{$^{1,3}$}} \Email{ege.oezsoy@tum.de}\\
\addr $^{1}$ Computer Aided Medical Procedures, TUM School of Computation, Information and Technology, Technical University of Munich, Boltzmannstr. 3, 85748 Garching, Germany.\\
\addr $^{3}$ Munich Center for Machine Learning (MCML), Boltzmannstrasse 3, 85748 Garching, Germany. \AND
\Name{Benjamin Busam\nametag{$^{1}$}} \Email{b.busam@tum.de}\\
\Name{Benedikt Wiestler\nametag{$^{2}$}} \Email{b.wiestler@tum.de}\\
\addr $^{2}$ AI for Image-Guided Diagnosis and Therapy, TUM School of Medicine and Health, Technical University of Munich, Ismaninger Str. 22, 81675 Munich, Germany. \AND
\Name{Nassir Navab\nametag{$^{1}$}} \Email{nassir.navab@tum.de}\\
\Name{Matthias Keicher\nametag{$^{1}$}} \Email{matthias.keicher@tum.de}
}

\begin{document}
\maketitle

\begin{abstract}
Conversational AI tools for generating and discussing accurate radiology reports could transform radiology by enabling collaborative, human-in-the-loop diagnostic processes, saving time and enhancing report quality. While, to this end, Large Vision-Language Models hold promise, current methods lack clinical correctness or are single-task models without conversational abilities. We propose a novel architecture and dataset to address these limitations. First, we propose a secondary image branch, explicitly focusing on structured clinical findings, improving the clinical correctness score by 13.3\%. Second, we propose a catastrophic forgetting mitigation strategy and instruct dataset with variable dialog-based tasks, to enable our model to handle a multitude of different queries. RaDialog marks a foundational step toward clinical dialog systems, outperforming existing medical LVLMs by 15.0\% in clinical correctness in report generation, 23.4\% in interactive report correction, and is preferred by radiologists in 84.0\% of cases over a comparative method. Our model and dataset are publicly available (\href{https://github.com/ChantalMP/RaDialog}{https://github.com/ChantalMP/RaDialog}, \href{https://physionet.org/content/radialog-instruct-dataset/1.1.0/}{https://physionet.org/content/radialog-instruct-dataset/1.1.0/}).

\end{abstract}


\section{Introduction}
\begin{figure*}[tb]
\centerline{\includegraphics[width=1.0\textwidth]{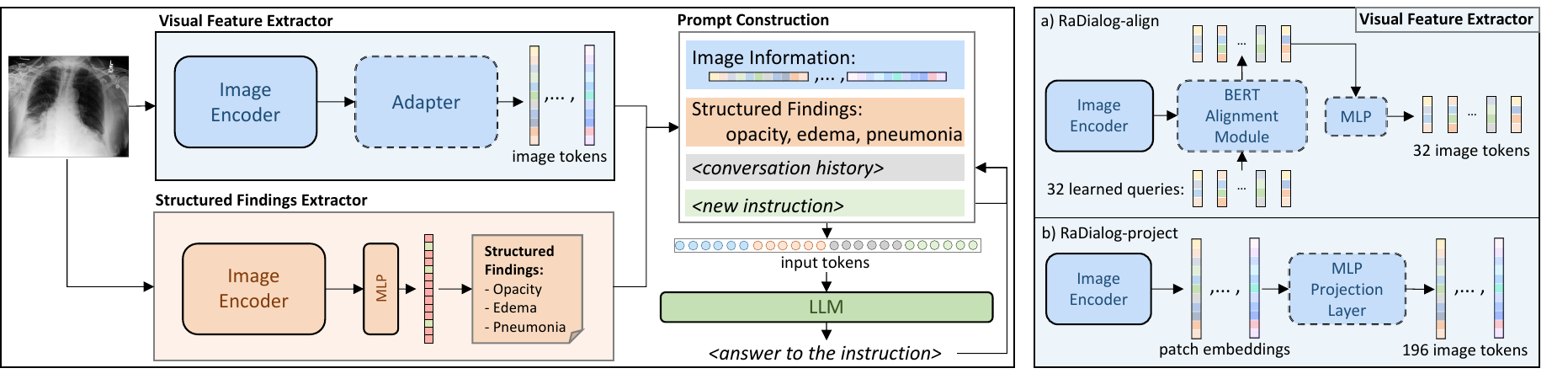}}
\caption{Pipeline overview: The Image Encoder extracts X-ray features and transforms them via adapter module a or b. The Structured Findings Extractor extracts high-level findings. Both outputs are integrated during Prompt Construction with conversation history and task-specific instructions to query the LLM. The predicted answer are added to the conversation history.}
\label{fig:model}
\end{figure*}
Radiology is crucial for clinical decision-making, with radiology reports serving as primary communication channel between radiologists and other clinicians, especially in the context of chest X-ray examinations, which are pivotal for identifying thoracic diseases~\cite{johnson2019mimic}. The rising volume of imaging exams underscores the need for automated report generation, which promises to simplify reporting and support radiologists~\cite{kaur2022methods}. Beyond mere report generation, dialog-based assistance holds potential for a more collaborative diagnostic process between radiologists and AI-based tools.\\
However, while state-of-the-art methods for radiology report generation produce coherent reports~\cite{wang2022inclusive,yang2022knowledge,hou-etal-2023-organ,wang2023metransformer,huang2023kiut,li2023dynamic}, they can struggle with factual correctness, and as single-task models, they are constrained to report generation as their only function.\\
Recent advances in large language models (LLMs) have demonstrated versatility across many tasks, including healthcare applications like medical exams and conversational diagnosis~\cite{touvron2023llama,vicuna2023,openai2023gpt4,singhal2023large,li2023chatdoctor,zhao2024chatcad+}. The development of large vision-language models (LVLMs) aims to equip these powerful LLMs with image understanding~\cite{li2023blip,li2024llava}. While several previous works specifically focus on medical imaging~\cite{tu2024towards,moor2023med,li2024llavamed,wu2023radfm,hyland2023maira,chen2024chexagent,thawkar2023xraygpt}, they are often limited to visual question-answering or single-step reporting tasks and lack robust interactive capabilities or clinical correctness. In contrast, RaDialog not only improves the accuracy of clinical report generation, but aims to enhance the radiology workflow by supporting dialog-based assistance for tasks such as report drafting, quick clarifications, collaborative insights, and reducing mental load for routine tasks.\\
Addressing key limitations of current methods, we propose RaDialog, a collaborative radiology assistant focusing on automated report generation and auxiliary interactive downstream tasks for chest X-rays. Our key contributions include:\\
\begin{itemize}[noitemsep, topsep=0pt]
    \item A novel dual-branch architecture that, inspired by structured reporting~\cite{pellegrini2023rad, keicher2023flexr}, incorporates a secondary visual feature extraction branch to focus on structured clinical findings, leading to 13.3\% improvement in clinical correctness score.
    \item A variable instruct training setup to enable dialog-based human-AI collaboration and combat the issue of catastrophic forgetting in LLM fine-tuning. Specifically, we design a semi-automatically labeled, image-grounded, interactive instruct dataset for X-Ray understanding, which we make publicly available.
    \item A context dropping augmentation, which randomly omits textual information in the conversation, requiring the model to consider the image information for all tasks.
    \item Demonstrated performance gains in both report generation and interactive tasks, outperforming XRayGPT~\cite{thawkar2023xraygpt} in direct comparisons, being preferred by radiologists in 84.0\% of cases.
\end{itemize}
By addressing these critical aspects, RaDialog represents a significant step forward in the development of clinical dialog systems for radiology. Our work paves the way for more accurate, versatile, and user-friendly AI assistants in medical imaging.

\section{Methodology}
\label{sec:meth}
\label{sec:exp}
RaDialog leverages Large Language Models (LLMs) and visual feature extraction techniques to address the complexities of medical imaging diagnostics, particularly focusing on chest X-rays. In this section, we present our model, training, and instruct dataset.

\subsection{Model and Training}
Our architecture, visualized in Fig.~\ref{fig:model}, consists of four main components: a Visual Feature Extractor, extracting image embeddings and aligning them to the text space; a Structured Findings Extractor to capture the presence of core findings; a Prompt Construction Module; and a Large Language Model (LLM), which outputs a response given image and instruction.\\
\textbf{Visual Feature Extractor} Given a chest X-ray image $x$, we first extract patch-wise image embeddings $x' \in \mathbb{R}^{P \times D_i}$ using a domain-specific X-ray encoder, where $P$ is the number of patches and $D_i$ is the dimensionality of each patch embedding. These patch-based features are passed to an adapter module, transforming them into $N$ embedded tokens $h \in \mathbb{R}^{N \times D_l}$, where $D_l$ is the dimension of the LLM tokens. For this adapter, we propose two variants, RaDialog-align and RaDialog-project, as depicted in Fig.~\ref{fig:model} a) and b). RaDialog-align, inspired by the architecture of BLIP-2~\cite{li2023blip}, uses a pre-trained BERT~\cite{devlin2018bert} model as alignment module, which is fine-tuned to embed the image information into $N=32$ tokens $h \in \mathbb{R}^{N \times D_q}$, given N learned query embeddings $q \in \mathbb{R}^{N \times D_q}$ as well as the output $x'$ of the image encoder. These tokens are then projected by an MLP to retrieve $N=32$ LLM input tokens. The alignment module is trained using three distinct objectives: an X-ray-report contrastive loss, a cross-entropy loss for image-report matching, and a language modeling loss for image-grounded report generation. This module is trained in a separate stage and remains frozen during the subsequent training of the large language model (LLM). RaDialog-project follows the image-to-text projection proposed in LLaVA~\cite{li2024llava}. Here, the patch features $x' \in \mathbb{R}^{P \times D_i}$ extracted by the image encoder are directly projected to $N=196$  language model tokens using an MLP as a projector to get the LLM input tokens $h = g(x')$ with $ h \in \mathbb{R}^{N \times D_l}$. The image encoder and adapter are trained jointly with the LLM without the need for a pre-training step.\\
\textbf{Structured Findings Extractor}
In addition to direct image features, in our secondary image branch, we build a structured representation of the main clinical findings in the image $x$. This enables our model to generate a free-text report that is aligned with these explicit findings, enhancing the controllability of the output and improving the clinical efficacy of our model. The Structured Findings Extractor consists of a CLIP vision encoder, initialized with pre-trained domain-specific weights, followed by a linear classification head and is separately trained for multi-label classification. The classification output is converted into text as a comma-separated list of all positive findings. The Structured Findings Extractor is trained using a log-weighted binary cross-entropy loss to address class imbalance.\\
\textbf{Prompt Construction}
Given $h = \{h_j\}_{j=1}^{N}, h_j \in \mathbb{R}^{D_l}$, the set of image tokens obtained from the Image Encoder; $S$, the description of structured findings; $H$, the conversation history; and $I$ the instruction, the LLM input prompt is constructed. The structured findings, conversation history, and instruction are embedded by the  LLM embedding layer $e(\cdot)$ into $e(S)$, $e(H)$, and $e(I)$. The final embedded prompt $P$ is constructed as concatenation of $\text(h, e(S), e(H), e(I))$. This prompt effectively leverages the strengths of both the encoding and structured information about the image, while considering the conversation context.\\
\textbf{Language Model}
Finally, the LLM processes the prompt $P$ and produces an instruction-specific response. Since the training data of generalist LLMs consists of limited medical information, we fine-tune our vision-language model on radiology reports and instructions using cross-entropy loss. This fine-tuning enhances both its medical knowledge and aligns its writing style with that of radiologists. Additionally, this fine-tuning trains the LLM to work effectively with image features and structured finding labels. For adapting the LLM, we use the parameter-efficient fine-tuning technique LoRA (Low-Rank Adaptation)~\cite{hu2021lora}, allowing domain adaptation with limited computational resources.

\subsection{Instruct Dataset}
Training solely on image-report pairs causes catastrophic forgetting in the LLM, reducing its ability to perform tasks beyond report generation. To address this, we design a diverse instruct dataset of $\approx$580k samples spanning ten tasks, each with ten prompt variations. It includes two types of tasks. Type 1 builds upon existing datasets~\cite{johnson2019mimic,kayser2022explaining,pellegrini2023rad} for report generation, impression generation, CheXpert QA, Rad-ReStruct QA, natural language explanations, and view classification. Type 2 comprises replay tasks, where we, inspired by continual learning, generate pseudo-ground truth with a non-fine-tuned LLM for correction, summarization, easy language, and region QA. While imperfect, this pseudo-ground truth mitigates catastrophic forgetting by replaying general language tasks during domain-specific training. Each sample includes an image, an instruction, and the corresponding ground truth. Tasks involving reports, like summarization, integrate the report generation instruction and ground truth report in the conversation history. More details and prompt examples for all tasks are shown in appendix \ref{ap:dataset}.\looseness=-1\\
\textbf{Context Dropping Augmentation}
For tasks that can be performed with only the image, and no report, as input, including Findings QA, Region QA, and View Classification, we propose to augment the information available to the model. Context Dropping systematically imitates potential inaccuracies in the initial report by varying the availability of textual input and prevents over-reliance on the text modality, which is generally easier to interpret for the model. By training the model under different levels of textual availability, we promote feature extraction from both visual and textual information, allowing the model to develop stronger visual reasoning capabilities when answering follow-up questions even when textual context is imperfect. We specify three configurations: $c=\{\text{\textit{full}},\textit{none},\textit{partial}\}$. In the \textit{full} mode, we keep the entire report $R$, while in the \textit{none} mode, the report and structured findings $S$ are dropped, so the model must rely on the visual encoding $h$ only.  In the \textit{partial} mode, half of the sentences in the report $R$ are randomly dropped. Let $h = \{h_ij\}_{j=1}^{N}$ be the set of image tokens, $S$ the structured findings, $H$ the conversation history including the ground truth report, and $I$ the instruction. The input prompt embedding $P$ is constructed based on the chosen configuration:
\begin{equation}
    P = 
    \begin{cases}
        \text{concat}(h, e(S), e(H), e(I)) & \text{if } c = \text{full} \\
        \text{concat}(h, e(I)) & \text{if } c = \text{none} \\
        \text{concat}(h, e(S), e(H'), e(I)) & \text{if } c = \text{partial}
    \end{cases}
\end{equation}
where $c \sim \mathcal{U}(\{\text{full}, \text{none}, \text{partial}\})$ is randomly chosen, and $H’$ is the history with $R$ replaced by $R_{partial}$, simulating incomplete report scenarios. This augmentation technique emphasizes the model’s focus on images rather than relying on the report content, enhancing its robustness and accuracy in scenarios with varying levels of report correctness.

\section{Experimental Setup}
\begin{figure}[tb]
  \centering
  \includegraphics[width=1.0\textwidth]{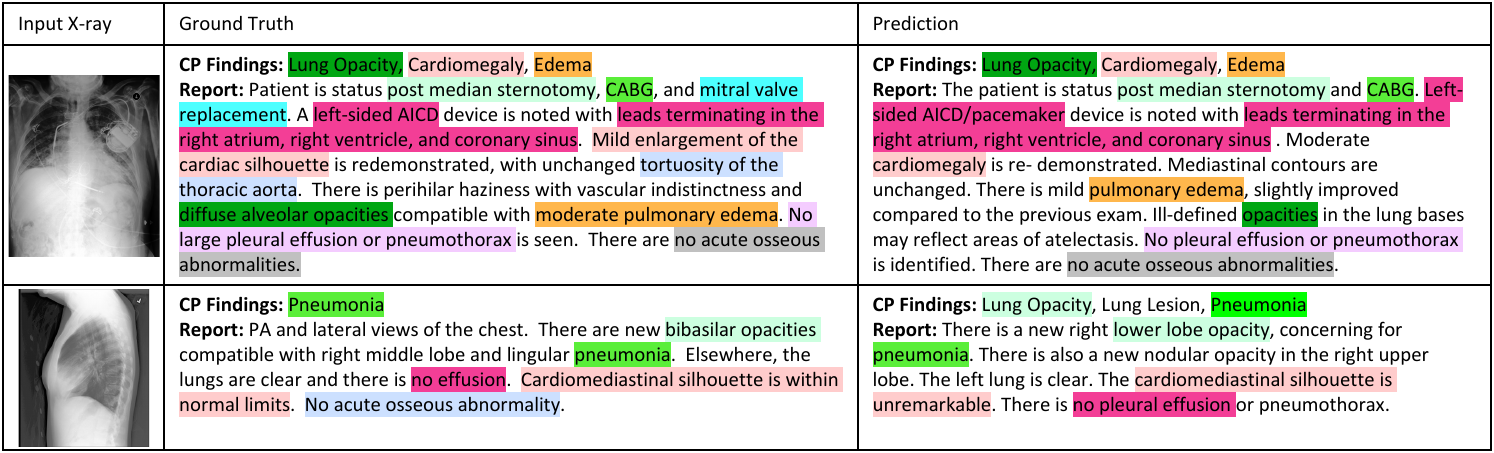}
   \caption{Qualitative report generation results of RaDialog\textsubscript{project} (top) and RaDialog\textsubscript{align} (bottom). Colors indicate matching findings in ground truth and prediction.}
   \label{fig:report_ex}
\end{figure}
We use the official splits of the widely used MIMIC-CXR~\cite{johnson2019mimic} dataset, comprising 377,110 chest X-rays and associated reports. Following prior work~\cite{chen2020generating,miura-etal-2021-improving,tanida2023interactive}, we predict the findings section of the reports and exclude samples with an empty findings section. For out-of-distribution evaluation, we use the test split of  IU-Xray~\cite{iu-xray}. We evaluate two model types, one trained only on report generation (RaDialog\textsubscript{rep}) and one trained on our instruct dataset (RaDialog\textsubscript{ins}), including both report generation and interactive downstream tasks.\\ 
We evaluate clinical efficacy (CE) as macro F1 over the 14 CheXbert labels~\cite{smit2020chexbert}, embedding-based text similarity (BertScore~\cite{zhang2019bertscore}), and standard Natural Language Generation (NLG) metrics (BLEU~\cite{papineni2002bleu}, ROUGE~\cite{lin-2004-rouge}, and METEOR~\cite{lavie2009meteor}). While conventional NLG metrics are not ideal for assessing the clinical correctness of radiology reports~\cite{pino2021clinically,yu2023evaluating,pellegrini2023rad}, they are included for completeness. More implementation Details are provided in \ref{ap:imp}.

\section{Results and Discussion}
\paragraph{Radiology report generation} We evaluate RaDialog on radiology report generation and compare it to recent LVLM-based methods and foundation models on MIMIC-CXR and IU-Xray in Tab.~\ref{tab:lvlms}. RaDialog outperforms all other methods significantly in clinical efficacy (CE) and BertScore (BS). The out-of-distribution (OOD) evaluation on IU-Xray, on which neither RaDialog nor most of the compared methods were trained, further showcases RaDialogs' benefit over prior LVLMs in X-ray report generation. Fig.~\ref{fig:report_ex} shows qualitative report generation results on a frontal and lateral chest X-ray. The color coding was verified by three board-certified radiologists. It can be observed that both our model variants capture a majority of the findings. RaDialog achieves an inference speed of 112 tokens/second for RaDialog\textsubscript{align} and 223 tokens/second for RaDialog\textsubscript{project}, resulting in an average generation time for an entire report of 1.2 or 0.6 seconds respectively. This speed is more than sufficient for integration into a radiologist’s workflow, where report generation typically takes up to several minutes. The appendix includes additional comparisons to older, non-LVLM-based methods (\ref{ap:rg}) and closed-source models using the indication as input (\ref{ap:ind}), reinforcing RaDialog’s strong performance.\\

\paragraph{Ablation of Architectural Components}
\label{sec:arch}
Table~\ref{tab:input} presents an ablation study evaluating the impact of fine-tuning the LLM and incorporating structured and visual image information. Comparing the first two rows, fine-tuning the LLM significantly improves performance, highlighting the importance of domain-specific adaptation. Furthermore, the results from rows two to four demonstrate that both structured and visual inputs contribute to performance gains, with the best results achieved when both modalities are combined.

\begin{table}[tb]
\small
  \centering
  \caption{Comparison of RaDialog to recent medical LVLMs. FT denotes if the model was fine-tuned on the respective dataset. $^a$trained in multi-view setting, but we evaluate with a single view. $\dagger$ re-computed results for MIMIC-CXR findings section}
  \begin{tabular}{lccccc|ccccc}
    \hline
    &\multicolumn{5}{c|}{MIMIC-CXR} & \multicolumn{5}{c}{IU-Xray (OOD)} \\
    \cline{1-11}
    Method & FT & CE & BS & B-4 & R-L & FT & CE & BS & B-4 & R-L \\
    \hline
    LLaVA-Med & $\times$ & \textit{10.7} & \textit{0.19} & \textit{1.1} & \textit{15.1} & $\times$ & 5.0 & 0.20 & 1.1 & 15.8 \\
    Rad-FM & $\checkmark$ & 15.4 & 0.22 & 2.4 & 15.6 & $\times$ & 5.9 & 0.20 & 2.3 & 13.8 \\
    XrayGPT & $\checkmark$ & 19.3 & 0.33 & 5.4 & 22.0 & $\times$ & 9.9 & 0.39 & 5.3 & 25.7 \\
    LLM-CXR & $\checkmark$ & 21.1 & - & - & - & - & - & - & - & - \\ 
    CheXagent$^a$ & $\checkmark$ & 22.2 & 0.36 & 7.3 & 25.9 & $\checkmark$ & \textit{14.1} & \textit{0.51} & \textit{12.7} & \textit{34.6} \\ 
    R2GenGPT$\dagger$ & $\checkmark$ & 24.7 & 0.36 & \textbf{10.1} & \textbf{27.6} & - & - & - & - & - \\ 
    \hline
    RaDialog\textsubscript{align-rep} & $\checkmark$ & 39.4 & \textbf{0.40} & 9.5 & 27.1 & $\times$ & 22.6 & \textbf{0.47} & 10.2 & \textbf{31.0} \\
    RaDialog\textsubscript{align-ins} & $\checkmark$ & 38.6 & 0.39 & 9.7 & 27.0 & $\times$ & 22.9 & 0.46 & 9.7 & 30.2 \\
    RaDialog\textsubscript{project-rep} & $\checkmark$ & \textbf{39.7} & 0.36 & 8.8 & 25.6 & $\times$ & 23.0 & 0.45 & 8.3 & 29.6 \\ 
    RaDialog\textsubscript{project-ins} & $\checkmark$ & 39.2 & 0.37 & 9.4 & 26.7 & $\times$ & \textbf{23.1} & 0.45 & \textbf{11.0} & 30.4 \\
    \hline
  \end{tabular}
  \label{tab:lvlms}
\end{table}

\begin{table}\setlength\tabcolsep{5pt}
  \centering
    \caption{Ablations of architectural components: compares using a non-fine-tuned LLM (NF) and the effect of visual (V) and structured (S) input. RaDialog-SFE refers to the classification metrics of the Structured Finding Extractor (SFE) of RaDialog.}
  \begin{tabular}{@{}lcccccccc@{}}
    \hline
    Method & V & S & CE & BS & B-1 & B-4 & MTR & R-L \\
    \hline
    RaDialog-align-NF &  $\times$  & \checkmark& 35.8 & 0.20 & 5.5 & 0.4 & 4.7 & 11.7\\
    RaDialog-align-report &  $\times$  & \checkmark & 37.3 & 0.39 & 32.6 & 8.2 & 12.8 & 25.9\\
    RaDialog-align-report & \checkmark &  $\times$  & 26.1 & 0.39 & 31.3 & 9.0 & 13.0 & \textbf{27.1}\\
    RaDialog-align-report & \checkmark & \checkmark & \textbf{39.4} & \textbf{0.40} & \textbf{34.6} & \textbf{9.5} & \textbf{14.0} & \textbf{27.1}\\ \hline
    RaDialog-SFE & - &  -  & 31.7 & - & - & - & - & -\\
    \hline
  \end{tabular}
  \label{tab:input}
\end{table}

\begin{figure}[tb]
  \centering
  \includegraphics[width=1.0\linewidth]{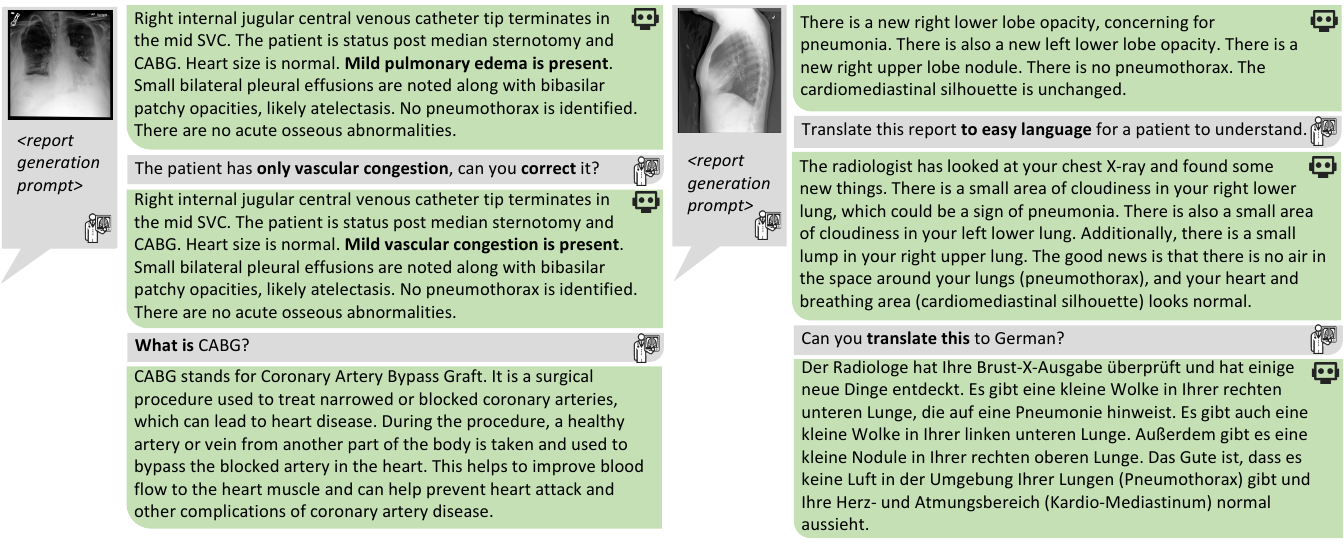}
   \caption{Qualitative conversation examples with RaDialog\textsubscript{project-ins} (left) and RaDialog\textsubscript{align-ins} (right), showing examples of correction, knowledge QA (zero-shot), easy language, and translation (zero-shot).}
   \label{fig:chat_ex}
\end{figure}

\begin{table}[tb]
  \centering
  \small
  \caption{Results on Impression Generation and View Classification.}
  \begin{tabular}{lcccccc}
    \hline
    \multirow{2}{*}{Method} & \multicolumn{4}{c}{Impression Generation} & \multicolumn{2}{c}{View Classification} \\
    \cline{2-5} \cline{6-7}
    & B-1 & B-4 & MTR & R-L & Accuracy & F1 \\
    \hline
    CheXagent & - & - & - & 40.3 & \textbf{97.5} & - \\
    RaDialog\textsubscript{project-rep} & 15.7 & 3.6 & 13.4 & 18.1 & 8.0 & 7.3 \\
    RaDialog\textsubscript{project-ins} & \textbf{40.0} & \textbf{19.5} & \textbf{19.9} & \textbf{45.8} & 97.1 & \textbf{95.9} \\
    \hline
  \end{tabular}
  \label{tab:results_impression_view}
\end{table}
\begin{table}[tb]
  \centering
  \small
  \caption{Findings QA and report correction results. "Initial", "Corrected", and "$\Delta$" represent CE scores before and after correction, and the resulting improvement.}
  \begin{tabular}{lcccccc|ccc}
    \hline
    \multirow{2}{*}{Method} & \multicolumn{6}{c|}{Findings QA} & \multicolumn{3}{c}{Report Correction} \\
    \cline{2-7} \cline{8-10}
    & \multicolumn{3}{c}{Binary Mode} & \multicolumn{3}{c|}{Complete Mode} & Initial & Corrected & $\Delta$ \\
    \cline{2-4} \cline{5-7}
    & F1 & Prec & Rec & F1 & Prec & Rec & & & \\
    \hline
    XrayGPT & 20.6 & 15.4 & 42.5 & - & - & - & 19.3 & 29.3 & 10.0 \\
    RaDialog\textsubscript{align-rep} & 1.8 & 17.3 & 7.5 & 9.8 & 16.0 & 12.5 & 39.4 & 49.9 & 10.5 \\
    RaDialog\textsubscript{align-ins} & \textbf{39.7} & 37.5 & \textbf{43.5} & \textbf{40.3} & \textbf{39.9} & \textbf{42.0} & 38.6 & 71.7 & 33.1 \\
    RaDialog\textsubscript{project-rep} & 29.3 & 33.0 & 30.0 & 16.2 & 28.0 & 28.1 & \textbf{39.7} & 64.7 & 25.0 \\
    RaDialog\textsubscript{project-ins} & 36.1 & \textbf{38.0} & 38.7 & 40.1 & 39.6 & \textbf{42.0} & 39.2 & \textbf{72.6} & \textbf{33.4} \\
    \hline
  \end{tabular}
  \label{tab:findingsqa_corr}
\end{table}
\paragraph{Interactive Downstream Tasks} Apart from report generation, we further evaluate our model on different interactive downstream tasks.\\
\textbf{Impression Generation:} Given the ground truth findings section, we generate the impression. Tab.~\ref{tab:results_impression_view} shows that our instruct training is crucial for this, outperforming the report-only model and CheXagent~\cite{chen2024chexagent} significantly.\\
\textbf{Report Correction:} For a quantitative evaluation, we generate correction prompts for the entire MIMIC-CXR test set, asking to correct all incorrect pathologies found by the CheXbert labeler~\cite{smit2020chexbert} in the initial predictions. Tab.~\ref{tab:findingsqa_corr} shows the improvement through correction, indicating that report correction leads to an improvement of the report of around 33\%, which is significantly higher than for our report-only models (10-25\%) and XRayGPT (10\%), the only other report generation method allowing interactive prompting.\looseness=-1\\
\textbf{Finding Prediction:} We ask the model to predict the main CheXpert findings for an image in either ``binary'' or ``complete'' mode. For the binary task, we ask for a single finding and check if the answer contains ``yes'' or ``no''. For the complete prediction, we ask for a list of all findings and check for all occurrences of the 14 CheXpert labels. The results in Tab.~\ref{tab:findingsqa_corr} indicate that the report-only models fall short on these tasks. In contrast, both our instruct models and XrayGPT show better performance, where our instruct models exhibit significantly superior results, emphasizing its high clinical correctness.\\
\textbf{Rad-ReStruct QA:} We evaluate our method on the Rad-ReStruct~\cite{pellegrini2023rad} dataset, a visual question-answering benchmark designed to populate structured radiology reports by answering hierarchical questions. Following the dataset’s proposed procedure, we iteratively construct a conversation by asking the model about findings and their attributes, incorporating prior questions, answers, and dataset-provided answer options into the prompt. Our instruct model achieves an F1 score of 29.5, a precision of 61.8, and a recall of 41.6, surpassing the recall of the specialized hi-VQA model (F1: 32.0, precision: 64.6, recall: 33.3) while maintaining competitive precision and F1 scores. We see a clear improvement compared to our report-only model (F1: 28.7, precision: 98.1, recall: 28.8). While the report-only model has a similar F1 score, this is mainly caused by the high precision, which is caused by the high majority of negative answers in the dataset, while the model almost always predicts a negative answer. The good results of RaDialog\textsubscript{ins}  opena path towards using specialized LVLMs for structured reporting of detailed findings.\\
\textbf{View Classification:} We ask the model given only an image, from which view this image was taken. We follow the experiment setup in CheXagent~\cite{chen2024chexagent}, and reach an almost perfect score in the task (see Tab.~\ref{tab:results_impression_view}, again showing RaDialog's ability to ground conversation answers on the image.\\
\textbf{Qualitative Conversation Results} Fig.~\ref{fig:chat_ex} shows conversation examples covering multiple tasks. Some of the tasks, such as correction and easy language, were part of our instruct dataset, while others, such as translation to another language and knowledge questions, were not seen during training. This shows how our training adapted the model to radiology-specific tasks while maintaining general capabilities of the used LLM. \\
Additionally, a board-certified radiologist compared both the report generation and conversational performance of RaDialog\textsubscript{align-ins} to XrayGPT using 50 randomly selected X-rays from the MIMIC-CXR test set. For each image, we asked both models to write a report and perform one of these conversational follow-up tasks: report correction, easy language, binary findings QA, summarization, translation to German, knowledge QA (e.g. ``What is an edema?'') and recommendation of follow-up diagnostics or treatment. For each image, reports were generated by both RaDialog\textsubscript{align-ins} and XrayGPT. The board-certified radiologist was presented with the original X-ray alongside the two generated reports, displayed in a randomized order to prevent bias. Without knowing the source of each report, the radiologist completed a structured three-question questionnaire per case, assessing their preference with regards to (1) report generation quality in terms of accuracy, completeness, and coherence, (2) conversational performance based on contextual relevance and correctness, and (3) overall.
The radiologist consistently favored RaDialog across all three evaluation criteria with a preference rate of 71\% in Conversational Performance and 84\% in Report Generation as well as Overall. This highlights the advantages of our approach, including the integration of structured findings, fine-tuning of the LLM with image data, and employing instruct training with replay examples, while X-rayGPT fine-tunes on text-only conversations in advance, keeping image encoder and LLM are frozen during alignment.\\
\textbf{Comparison of RaDialog\textsubscript{align} and RaDialog\textsubscript{project}}
Even though the architecture and training procedure RaDialog\textsubscript{project} is more straightforward, both adapter types have a very similar performance in report generation and the evaluated downstream tasks. However, considering qualitative conversational performance of both models it can be observed the \textit{project} variant is more robust to zero-shot tasks, we did not explicitly train on, such as knowledge questions or treatment suggestions. An example is provided in appendix \ref{ap:qual}.

\noindent\textbf{Limitations}
We performed an additional failure case analysis on 50 RaDialog predictions on the MIMIC-CXR test set. We observe that in general, Radialog performs well in identifying major radiological findings and the presence of support devices, showcasing its ability to detect significant abnormalities and medical hardware. The most frequent errors are mischaracterizing less frequent findings, misjudging the severity of conditions, and occasionally introducing minor incorrect details such as the exact position of findings or devices. While not flawless, Radialog can provide valuable utility for handling simpler cases, providing initial drafts and enabling collaborative radiologist-AI reporting.

\section{Conclusion}
\label{sec:conc}
In this work, we introduced RaDialog, a novel large vision-language model for the generation of radiology reports and auxiliary interactive assistance. Besides accurate report generation abilities, our model can also engage in a dialog, answer follow-up questions, and incorporate feedback, enabling intuitive quality control through experts in the loop. By incorporating intermediate structured radiology findings in a secondary image branch, our method reaches state-of-the-art results in creating clinically accurate reports. Secondly, through our instruct training setup on our publicly available instruct dataset dialog-based assistance is enabled, by avoiding catastrophic forgetting and teaching domain-specific conversational tasks. Lastly, by augmenting our dataset with context dropping, we enforce attention on the input image throughout the entire conversation. We believe RaDialog represents a significant leap forward from static automated report generation to a more dynamic, collaborative tool that mirrors the interactive nature of clinical practice and encourages the community to explore more collaborative medical image understanding approaches.

\FloatBarrier

\clearpage  
\midlacknowledgments{This work was supported in part by the Federal Ministry of Education and Research of Germany (BMBF) under project DIVA (13GW0469C), the Bavarian Ministry of Economic Affairs, Regional Development and Energy (StMWi) under project ThoraXAI (DIK-2302-0002), and the German Research Foundation (DFG, grant 469106425 - NA 620/51-1).}

\bibliography{midl25_89}

\begin{thebibliography}{48}
\providecommand{\natexlab}[1]{#1}
\providecommand{\url}[1]{\texttt{#1}}
\expandafter\ifx\csname urlstyle\endcsname\relax
  \providecommand{\doi}[1]{doi: #1}\else
  \providecommand{\doi}{doi: \begingroup \urlstyle{rm}\Url}\fi

\bibitem[Achiam et~al.(2023)Achiam, Adler, Agarwal, Ahmad, Akkaya, Aleman, Almeida, Altenschmidt, Altman, Anadkat, et~al.]{openai2023gpt4}
Josh Achiam, Steven Adler, Sandhini Agarwal, Lama Ahmad, Ilge Akkaya, Florencia~Leoni Aleman, Diogo Almeida, Janko Altenschmidt, Sam Altman, Shyamal Anadkat, et~al.
\newblock Gpt-4 technical report.
\newblock \emph{arXiv}, 2023.

\bibitem[Bannur et~al.(2023)Bannur, Hyland, Liu, Perez-Garcia, Ilse, Castro, Boecking, Sharma, Bouzid, Thieme, et~al.]{bannur2023learning}
Shruthi Bannur, Stephanie Hyland, Qianchu Liu, Fernando Perez-Garcia, Maximilian Ilse, Daniel~C Castro, Benedikt Boecking, Harshita Sharma, Kenza Bouzid, Anja Thieme, et~al.
\newblock Learning to exploit temporal structure for biomedical vision-language processing.
\newblock In \emph{CVPR}, 2023.

\bibitem[Chen et~al.(2020)Chen, Song, Chang, and Wan]{chen2020generating}
Zhihong Chen, Yan Song, Tsung-Hui Chang, and Xiang Wan.
\newblock Generating radiology reports via memory-driven transformer.
\newblock In \emph{EMNLP}, 2020.

\bibitem[Chen et~al.(2024)Chen, Varma, Delbrouck, Paschali, Blankemeier, Van~Veen, Valanarasu, Youssef, Cohen, Reis, et~al.]{chen2024chexagent}
Zhihong Chen, Maya Varma, Jean-Benoit Delbrouck, Magdalini Paschali, Louis Blankemeier, Dave Van~Veen, Jeya Maria~Jose Valanarasu, Alaa Youssef, Joseph~Paul Cohen, Eduardo~Pontes Reis, et~al.
\newblock Chexagent: Towards a foundation model for chest x-ray interpretation.
\newblock \emph{arXiv}, 2024.

\bibitem[Chiang et~al.(2023)Chiang, Li, Lin, Sheng, Wu, Zhang, Zheng, Zhuang, Zhuang, Gonzalez, Stoica, and Xing]{vicuna2023}
Wei-Lin Chiang, Zhuohan Li, Zi~Lin, Ying Sheng, Zhanghao Wu, Hao Zhang, Lianmin Zheng, Siyuan Zhuang, Yonghao Zhuang, Joseph~E. Gonzalez, Ion Stoica, and Eric~P. Xing.
\newblock Vicuna: An open-source chatbot impressing gpt-4 with 90\%* chatgpt quality, 2023.

\bibitem[Demner-Fushman et~al.(2016)Demner-Fushman, Kohli, Rosenman, Shooshan, Rodriguez, Antani, Thoma, and McDonald]{iu-xray}
Dina Demner-Fushman, Marc~D Kohli, Marc~B Rosenman, Sonya~E Shooshan, Laritza Rodriguez, Sameer Antani, George~R Thoma, and Clement~J McDonald.
\newblock Preparing a collection of radiology examinations for distribution and retrieval.
\newblock \emph{JAMIA}, 2016.

\bibitem[Devlin et~al.(2018)Devlin, Chang, Lee, and Toutanova]{devlin2018bert}
Jacob Devlin, Ming-Wei Chang, Kenton Lee, and Kristina Toutanova.
\newblock Bert: Pre-training of deep bidirectional transformers for language understanding.
\newblock \emph{arXiv}, 2018.

\bibitem[Gu et~al.(2024{\natexlab{a}})Gu, Liu, Li, and Cai]{gu2024complex}
Tiancheng Gu, Dongnan Liu, Zhiyuan Li, and Weidong Cai.
\newblock Complex organ mask guided radiology report generation.
\newblock In \emph{Proceedings of the IEEE/CVF Winter Conference on Applications of Computer Vision}, pages 7995--8004, 2024{\natexlab{a}}.

\bibitem[Gu et~al.(2024{\natexlab{b}})Gu, Yang, An, Feng, Liu, and Cai]{gu2024orid}
Tiancheng Gu, Kaicheng Yang, Xiang An, Ziyong Feng, Dongnan Liu, and Weidong Cai.
\newblock Orid: Organ-regional information driven framework for radiology report generation.
\newblock \emph{arXiv preprint arXiv:2411.13025}, 2024{\natexlab{b}}.

\bibitem[Hou et~al.(2023)Hou, Xu, Cheng, Li, and Liu]{hou-etal-2023-organ}
Wenjun Hou, Kaishuai Xu, Yi~Cheng, Wenjie Li, and Jiang Liu.
\newblock {ORGAN}: Observation-guided radiology report generation via tree reasoning.
\newblock In \emph{ACL}, 2023.

\bibitem[Hu et~al.(2021)Hu, Shen, Wallis, Allen-Zhu, Li, Wang, Wang, and Chen]{hu2021lora}
Edward~J Hu, Yelong Shen, Phillip Wallis, Zeyuan Allen-Zhu, Yuanzhi Li, Shean Wang, Lu~Wang, and Weizhu Chen.
\newblock Lora: Low-rank adaptation of large language models.
\newblock \emph{arXiv}, 2021.

\bibitem[Huang et~al.(2023)Huang, Zhang, and Zhang]{huang2023kiut}
Zhongzhen Huang, Xiaofan Zhang, and Shaoting Zhang.
\newblock Kiut: Knowledge-injected u-transformer for radiology report generation.
\newblock In \emph{CVPR}, 2023.

\bibitem[Hyland et~al.(2023)Hyland, Bannur, Bouzid, Castro, Ranjit, Schwaighofer, P{\'e}rez-Garc{\'\i}a, Salvatelli, Srivastav, Thieme, et~al.]{hyland2023maira}
Stephanie~L Hyland, Shruthi Bannur, Kenza Bouzid, Daniel~C Castro, Mercy Ranjit, Anton Schwaighofer, Fernando P{\'e}rez-Garc{\'\i}a, Valentina Salvatelli, Shaury Srivastav, Anja Thieme, et~al.
\newblock Maira-1: A specialised large multimodal model for radiology report generation.
\newblock \emph{arXiv}, 2023.

\bibitem[Jeong et~al.(2023)Jeong, Tian, Li, Hartung, Behzadi, Calle, Osayande, Pohlen, Adithan, and Rajpurkar]{jeong2023multimodal}
Jaehwan Jeong, Katherine Tian, Andrew Li, Sina Hartung, Fardad Behzadi, Juan Calle, David Osayande, Michael Pohlen, Subathra Adithan, and Pranav Rajpurkar.
\newblock Multimodal image-text matching improves retrieval-based chest x-ray report generation.
\newblock \emph{arXiv}, 2023.

\bibitem[Johnson et~al.(2019)Johnson, Pollard, Berkowitz, Greenbaum, Lungren, Deng, Mark, and Horng]{johnson2019mimic}
Alistair~EW Johnson, Tom~J Pollard, Seth~J Berkowitz, Nathaniel~R Greenbaum, Matthew~P Lungren, Chih-ying Deng, Roger~G Mark, and Steven Horng.
\newblock Mimic-cxr, a de-identified publicly available database of chest radiographs with free-text reports.
\newblock \emph{Scientific data}, 2019.

\bibitem[Kaur et~al.(2022)Kaur, Mittal, and Singh]{kaur2022methods}
Navdeep Kaur, Ajay Mittal, and Gurprem Singh.
\newblock Methods for automatic generation of radiological reports of chest radiographs: a comprehensive survey.
\newblock \emph{Multimed Tools Appl}, 2022.

\bibitem[Kayser et~al.(2022)Kayser, Emde, Camburu, Parsons, Papiez, and Lukasiewicz]{kayser2022explaining}
Maxime Kayser, Cornelius Emde, Oana-Maria Camburu, Guy Parsons, Bartlomiej Papiez, and Thomas Lukasiewicz.
\newblock Explaining chest x-ray pathologies in natural language.
\newblock In \emph{MICCAI}, 2022.

\bibitem[Keicher et~al.(2023)Keicher, Zaripova, Czempiel, Mach, Khakzar, and Navab]{keicher2023flexr}
Matthias Keicher, Kamilia Zaripova, Tobias Czempiel, Kristina Mach, Ashkan Khakzar, and Nassir Navab.
\newblock Flexr: Few-shot classification with language embeddings for structured reporting of chest x-rays.
\newblock In \emph{MIDL}, 2023.

\bibitem[Lavie and Denkowski(2009)]{lavie2009meteor}
Alon Lavie and Michael~J Denkowski.
\newblock The meteor metric for automatic evaluation of machine translation.
\newblock \emph{Machine translation}, 2009.

\bibitem[Li et~al.(2024)Li, Wong, Zhang, Usuyama, Liu, Yang, Naumann, Poon, and Gao]{li2024llavamed}
Chunyuan Li, Cliff Wong, Sheng Zhang, Naoto Usuyama, Haotian Liu, Jianwei Yang, Tristan Naumann, Hoifung Poon, and Jianfeng Gao.
\newblock Llava-med: Training a large language-and-vision assistant for biomedicine in one day.
\newblock \emph{NIPS}, 2024.

\bibitem[Li et~al.(2023{\natexlab{a}})Li, Li, Savarese, and Hoi]{li2023blip}
Junnan Li, Dongxu Li, Silvio Savarese, and Steven Hoi.
\newblock Blip-2: Bootstrapping language-image pre-training with frozen image encoders and large language models.
\newblock \emph{arXiv}, 2023{\natexlab{a}}.

\bibitem[Li et~al.(2023{\natexlab{b}})Li, Lin, Chen, Lin, Liang, and Chang]{li2023dynamic}
Mingjie Li, Bingqian Lin, Zicong Chen, Haokun Lin, Xiaodan Liang, and Xiaojun Chang.
\newblock Dynamic graph enhanced contrastive learning for chest x-ray report generation.
\newblock In \emph{CVPR}, 2023{\natexlab{b}}.

\bibitem[Li et~al.(2023{\natexlab{c}})Li, Li, Zhang, Dan, Jiang, and Zhang]{li2023chatdoctor}
Yunxiang Li, Zihan Li, Kai Zhang, Ruilong Dan, Steve Jiang, and You Zhang.
\newblock Chatdoctor: A medical chat model fine-tuned on a large language model meta-ai (llama) using medical domain knowledge.
\newblock \emph{Cureus}, 2023{\natexlab{c}}.

\bibitem[Lin(2004)]{lin-2004-rouge}
Chin-Yew Lin.
\newblock {ROUGE}: A package for automatic evaluation of summaries.
\newblock In \emph{Text Summarization Branches Out}. ACL, 2004.

\bibitem[Liu et~al.(2024)Liu, Li, Wu, and Lee]{li2024llava}
Haotian Liu, Chunyuan Li, Qingyang Wu, and Yong~Jae Lee.
\newblock Visual instruction tuning.
\newblock \emph{Advances in neural information processing systems}, 36, 2024.

\bibitem[Miura et~al.(2021)Miura, Zhang, Tsai, Langlotz, and Jurafsky]{miura-etal-2021-improving}
Yasuhide Miura, Yuhao Zhang, Emily Tsai, Curtis Langlotz, and Dan Jurafsky.
\newblock Improving factual completeness and consistency of image-to-text radiology report generation.
\newblock In \emph{NAACL}, 2021.

\bibitem[Moor et~al.(2023)Moor, Huang, Wu, Yasunaga, Dalmia, Leskovec, Zakka, Reis, and Rajpurkar]{moor2023med}
Michael Moor, Qian Huang, Shirley Wu, Michihiro Yasunaga, Yash Dalmia, Jure Leskovec, Cyril Zakka, Eduardo~Pontes Reis, and Pranav Rajpurkar.
\newblock Med-flamingo: a multimodal medical few-shot learner.
\newblock In \emph{ML4H}, 2023.

\bibitem[Nooralahzadeh et~al.(2021)Nooralahzadeh, Gonzalez, Frauenfelder, Fujimoto, and Krauthammer]{nooralahzadeh2021progressive}
Farhad Nooralahzadeh, Nicolas~Perez Gonzalez, Thomas Frauenfelder, Koji Fujimoto, and Michael Krauthammer.
\newblock Progressive transformer-based generation of radiology reports.
\newblock In \emph{EMNLP}, 2021.

\bibitem[Papineni et~al.(2002)Papineni, Roukos, Ward, and Zhu]{papineni2002bleu}
Kishore Papineni, Salim Roukos, Todd Ward, and Wei-Jing Zhu.
\newblock Bleu: a method for automatic evaluation of machine translation.
\newblock In \emph{ACL}, 2002.

\bibitem[Pellegrini et~al.(2023)Pellegrini, Keicher, {\"O}zsoy, and Navab]{pellegrini2023rad}
Chantal Pellegrini, Matthias Keicher, Ege {\"O}zsoy, and Nassir Navab.
\newblock Rad-restruct: A novel vqa benchmark and method for structured radiology reporting.
\newblock In \emph{MICCAI}. Springer, 2023.

\bibitem[Pino et~al.(2021)Pino, Parra, Besa, and Lagos]{pino2021clinically}
Pablo Pino, Denis Parra, Cecilia Besa, and Claudio Lagos.
\newblock Clinically correct report generation from chest x-rays using templates.
\newblock In \emph{MLMI 2021 at MICCAI 2021}. Springer, 2021.

\bibitem[Singhal et~al.(2023)Singhal, Azizi, Tu, Mahdavi, Wei, Chung, Scales, Tanwani, Cole-Lewis, Pfohl, et~al.]{singhal2023large}
Karan Singhal, Shekoofeh Azizi, Tao Tu, S~Sara Mahdavi, Jason Wei, Hyung~Won Chung, Nathan Scales, Ajay Tanwani, Heather Cole-Lewis, Stephen Pfohl, et~al.
\newblock Large language models encode clinical knowledge.
\newblock \emph{Nature}, 2023.

\bibitem[Smit et~al.(2020)Smit, Jain, Rajpurkar, Pareek, Ng, and Lungren]{smit2020chexbert}
Akshay Smit, Saahil Jain, Pranav Rajpurkar, Anuj Pareek, Andrew~Y Ng, and Matthew~P Lungren.
\newblock Chexbert: combining automatic labelers and expert annotations for accurate radiology report labeling using bert.
\newblock \emph{arXiv}, 2020.

\bibitem[Tanida et~al.(2023)Tanida, Müller, Kaissis, and Rueckert]{tanida2023interactive}
Tim Tanida, Philip Müller, Georgios Kaissis, and Daniel Rueckert.
\newblock Interactive and explainable region-guided radiology report generation.
\newblock In \emph{CVPR}, 2023.

\bibitem[Thawkar et~al.(2023)Thawkar, Shaker, Mullappilly, Cholakkal, Anwer, Khan, Laaksonen, and Khan]{thawkar2023xraygpt}
Omkar Thawkar, Abdelrahman Shaker, Sahal~Shaji Mullappilly, Hisham Cholakkal, Rao~Muhammad Anwer, Salman Khan, Jorma Laaksonen, and Fahad~Shahbaz Khan.
\newblock Xraygpt: Chest radiographs summarization using medical vision-language models.
\newblock \emph{arXiv}, 2023.

\bibitem[Touvron et~al.(2023)Touvron, Lavril, Izacard, Martinet, Lachaux, Lacroix, Rozi{\`e}re, Goyal, Hambro, Azhar, et~al.]{touvron2023llama}
Hugo Touvron, Thibaut Lavril, Gautier Izacard, Xavier Martinet, Marie-Anne Lachaux, Timoth{\'e}e Lacroix, Baptiste Rozi{\`e}re, Naman Goyal, Eric Hambro, Faisal Azhar, et~al.
\newblock Llama: Open and efficient foundation language models.
\newblock \emph{arXiv}, 2023.

\bibitem[Tu et~al.(2024)Tu, Azizi, Driess, Schaekermann, Amin, Chang, Carroll, Lau, Tanno, Ktena, et~al.]{tu2024towards}
Tao Tu, Shekoofeh Azizi, Danny Driess, Mike Schaekermann, Mohamed Amin, Pi-Chuan Chang, Andrew Carroll, Charles Lau, Ryutaro Tanno, Ira Ktena, et~al.
\newblock Towards generalist biomedical ai.
\newblock \emph{NEJM AI}, 2024.

\bibitem[Wang et~al.(2025)Wang, Teng, Zhang, Yang, Wang, Yi, Zhang, Wang, Tavares, and Xu]{wang2025hkrg}
Bo~Wang, Peihong Teng, Hongda Zhang, Feiyang Yang, Zeyu Wang, Xingcheng Yi, Tianyang Zhang, Chunhui Wang, Adriano~Jose Tavares, and Hao Xu.
\newblock Hkrg: Hierarchical knowledge integration for radiology report generation.
\newblock \emph{Expert Systems with Applications}, 271:\penalty0 126622, 2025.

\bibitem[Wang et~al.(2022)Wang, Ning, Lu, Wei, Zheng, and Chen]{wang2022inclusive}
Lin Wang, Munan Ning, Donghuan Lu, Dong Wei, Yefeng Zheng, and Jie Chen.
\newblock An inclusive task-aware framework for radiology report generation.
\newblock In \emph{MICCAI}. Springer, 2022.

\bibitem[Wang et~al.(2023)Wang, Liu, Wang, and Zhou]{wang2023metransformer}
Zhanyu Wang, Lingqiao Liu, Lei Wang, and Luping Zhou.
\newblock Metransformer: Radiology report generation by transformer with multiple learnable expert tokens.
\newblock In \emph{CVPR}, 2023.

\bibitem[Wu et~al.(2023)Wu, Zhang, Zhang, Wang, and Xie]{wu2023radfm}
Chaoyi Wu, Xiaoman Zhang, Ya~Zhang, Yanfeng Wang, and Weidi Xie.
\newblock Towards generalist foundation model for radiology.
\newblock \emph{arXiv}, 2023.

\bibitem[Xiao et~al.(2024)Xiao, Shi, Liu, Wang, and Bai]{xiao2024radiology}
Ting Xiao, Lei Shi, Peng Liu, Zhe Wang, and Chenjia Bai.
\newblock Radiology report generation via multi-objective preference optimization.
\newblock \emph{arXiv preprint arXiv:2412.08901}, 2024.

\bibitem[Yan et~al.(2021)Yan, He, Lu, Du, Chang, Gentili, McAuley, and Hsu]{yan2021weakly}
An~Yan, Zexue He, Xing Lu, Jiang Du, Eric Chang, Amilcare Gentili, Julian McAuley, and Chun-nan Hsu.
\newblock Weakly supervised contrastive learning for chest x-ray report generation.
\newblock In \emph{EMNLP}, 2021.

\bibitem[Yang et~al.(2022)Yang, Wu, Ge, Zhou, and Xiao]{yang2022knowledge}
Shuxin Yang, Xian Wu, Shen Ge, S~Kevin Zhou, and Li~Xiao.
\newblock Knowledge matters: Chest radiology report generation with general and specific knowledge.
\newblock \emph{MedIA}, 2022.

\bibitem[Yang et~al.(2023)Yang, Wu, Ge, Zheng, Zhou, and Xiao]{yang2023radiology}
Shuxin Yang, Xian Wu, Shen Ge, Zhuozhao Zheng, S~Kevin Zhou, and Li~Xiao.
\newblock Radiology report generation with a learned knowledge base and multi-modal alignment.
\newblock \emph{Medical Image Analysis}, 86:\penalty0 102798, 2023.

\bibitem[Yu et~al.(2023)Yu, Endo, Krishnan, Pan, Tsai, Reis, Fonseca, Lee, Abad, Ng, et~al.]{yu2023evaluating}
Feiyang Yu, Mark Endo, Rayan Krishnan, Ian Pan, Andy Tsai, Eduardo~Pontes Reis, Eduardo Kaiser Ururahy~Nunes Fonseca, Henrique Min~Ho Lee, Zahra Shakeri~Hossein Abad, Andrew~Y Ng, et~al.
\newblock Evaluating progress in automatic chest x-ray radiology report generation.
\newblock \emph{Patterns}, 2023.

\bibitem[Zhang et~al.(2019)Zhang, Kishore, Wu, Weinberger, and Artzi]{zhang2019bertscore}
Tianyi Zhang, Varsha Kishore, Felix Wu, Kilian~Q Weinberger, and Yoav Artzi.
\newblock Bertscore: Evaluating text generation with bert.
\newblock In \emph{ICLR}, 2019.

\bibitem[Zhao et~al.(2024)Zhao, Wang, Gu, Zhu, Mei, Zhuang, Cui, Wang, and Shen]{zhao2024chatcad+}
Zihao Zhao, Sheng Wang, Jinchen Gu, Yitao Zhu, Lanzhuju Mei, Zixu Zhuang, Zhiming Cui, Qian Wang, and Dinggang Shen.
\newblock Chatcad+: Towards a universal and reliable interactive cad using llms.
\newblock \emph{IEEE Transactions on Medical Imaging}, 2024.

\end{thebibliography}

\appendix

\section{Additional report generation results }
\label{ap:rg}

We evaluate RaDialog on radiology report generation and compare other traditional single-task report generation methods evaluated on the findings sections of the official MIMIC-CXR test set in Tab.~\ref{tab:rg}. We do not include methods that use incomparable ground truth in terms of used test split~\cite{tanida2023interactive}, CE score definition~\cite{hou-etal-2023-organ} or reports~\cite{yang2022knowledge}. RaDialog outperforms all prior works in the clinical efficacy metric, demonstrating our model's ability to infer a correct clinical diagnosis. We also outperform previous methods in the BertScore, indicating that RaDialog often predicts correct content even if the formulation differs. We hypothesize that while an LLM understands context and semantics more deeply, a smaller model trained only on a specific dataset may mirror the dataset's exact wording more closely, resulting in higher NLG scores without necessarily improving clinical correctness.

\begin{table}
\centering
\caption{Performance comparison of RaDialog to existing methods on MIMIC-CXR~\cite{johnson2019mimic} with respect to CE and NLG metrics.}
\label{tab:rg}
\begin{tabular}{@{}lclccc@{}}
  \hline
  Method & CE & BS & B-4 & MTR & R-L \\
  \hline
  R2Gen~\cite{chen2020generating} & 27.6 & 0.27$^a$ & 10.3 & 14.2 & 27.7\\
  MDT+WCL~\cite{yan2021weakly} & 29.4 & 0.28$^a$ & 10.7 & 14.4 & 27.4 \\
  M\textsuperscript{2} Tr.~\cite{nooralahzadeh2021progressive} & 30.8 & 0.39$^a$ & 10.7 & 14.5 & 27.2\\
  ITA~\cite{wang2022inclusive} & 30.8 & - & 12.1 & 14.7 & 28.4 \\
  METransformer~\cite{wang2023metransformer} & 31.1 & - & \textbf{12.4} & 15.2 & \textbf{29.1}\\
  Kiut~\cite{huang2023kiut} & 32.1 & - & 11.3 & \textbf{16.0} & 28.5 \\
  M2KT~\cite{yang2023radiology} & 35.2 & - & 11.1 & - & 27.4 \\ 
  COMG~\cite{gu2024complex} & 34.5 & - & 10.4 & 13.7 & 27.9 \\ 
  HKRG~\cite{wang2025hkrg} & 33.9 & - & 14.3 & 16.7 & 31.0 \\ 
  ORID~\cite{gu2024orid} & 35.2 & - & 11.6 & 15.0 & 28.4 \\ 
  MPO~\cite{xiao2024radiology} & 35.3 & - & 13.9 & 16.2 & 30.9 \\

  \hline
  RaDialog-align-report & 39.4 & \textbf{0.40} & 9.5 & 14.0 & 26.7\\
  RaDialog-align-instruct & 38.6 & 0.39 & 9.7 & 13.6 & 27.0\\
  RaDialog-project-report & \textbf{39.7} & 0.36 & 8.8 & 14.4 & 25.6\\
  RaDialog-project-instruct & 39.2 & 0.37 & 9.4 & 14.2 & 26.7\\
  \hline
  \multicolumn{6}{p{230pt}}{$^a$ values reported from~\cite{jeong2023multimodal}} \\ 
\end{tabular}
\end{table}

\section{Details on LVLM Configurations}
Table \ref{tab:lvlms-features} provides a detailed comparison of the configurations of all evaluated LVLMs. RaDialog does not rely on a significantly larger or more powerful image encoder, nor does it process substantially more image tokens or utilize a larger LLM compared to other models. Instead, its key differentiating factors lie in the end-to-end fine-tuning of both the image and encoder and the LLM, the proposed dual-branch architecture and the task-specific training via the RaDialog-Instruct dataset, which are designed to enhance medical report generation and diagnostic conversational abilities.

\begin{table}[ht]
\small 
\centering
\caption{Comparison of configurations of RaDialog and other medical LVLMs. Abbreviations: img enc. = image encoder, tokens/img = image tokens per sample, LLM = large language model, ds = datasets.}
\begin{tabular}{p{2.2cm} p{2.3cm} p{1.1cm} p{1.4cm} p{2.7cm} p{2.8cm}}
    \hline
    \textbf{Method} & \textbf{Img enc.} & \textbf{LLM size} & \textbf{Tokens per img} & \textbf{Datasets} & \textbf{end-to-end training LLM/img enc.} \\ 
    \hline
    LLaVA-Med & CLIP-ViT & 7B & 576 & PMC-15M & \checkmark / $\times$ \\ 
    Rad-FM & 3D ViT & 13B & 32 & mix of 18 datasets & \checkmark / \checkmark \\  
    XrayGPT & MedClip  & 7B & 512 & MIMIC-CXR, IU-Xray & $\times$ / $\times$ \\ 
    LLM-CXR & VQ-GAN & 3B & 256 & MIMIC-CXR & \checkmark / $\times$ \\  
    CheXagent & EVA-CLIP-g & 7B & 128 & mix of 28 datasets & \checkmark / \checkmark \\ 
    R2GenGPT & Swin-T & 7B & 49 & MIMIC-CXR & $\times$ / \checkmark \\  
    \hline
    RaDialog\textsubscript{align} & BioVil-T & 7B & 32 & MIMIC-CXR, RaDialog-Instruct & \checkmark  / \checkmark\\
    RaDialog\textsubscript{project} & BioVil-T & 7B & 196 & MIMIC-CXR, RaDialog-Instruct & \checkmark  / \checkmark\\ 
    \hline
\end{tabular}
\label{tab:lvlms-features}
\end{table}
\section{Instruct Dataset Details}
\label{ap:dataset}

\subsection{Task Descriptions}

\textbf{Report Generation:} Produce the findings section of a radiology report given an X-ray. We use the MIMIC-CXR dataset~\cite{johnson2019mimic} as ground truth.

\textbf{Impression Generation:} Given the findings section of a radiology report, write the corresponding impression section. The ground truth impression sections are extracted from the MIMIC-CXR dataset~\cite{johnson2019mimic}.

\textbf{Findings QA:} Answer a question about the CheXpert labels by either listing all findings (complete) in the image or providing a yes/no answer about a specific finding (binary). We employ MIMIC-CXR CheXbert~\cite{smit2020chexbert} labels for supervision.

\textbf{Rad-ReStruct QA:} Answer detailed questions about the existence, location, and appearance of various chest X-ray findings in order to construct a structured report. As ground truth questions and answers, we use the samples from the Rad-ReStruct~\cite{pellegrini2023rad} dataset. As this dataset is highly imbalanced, we restrict our training data to include the same number of positive and negative samples for each question type.

\textbf{Region QA:} Answer a question about a specific region, such as the lungs, which can be binary or open-ended. The supervision signal is LLM-generated.

\textbf{Easy Language:} Reformulate the produced report into a simpler and more understandable language. The supervision signal is LLM-generated. 

\textbf{Summarization:} Summarize the report as bullet points or a short text. The supervision signal is LLM-generated.

\textbf{Correction:} Correct an error in the produced report. The training samples are generated by detecting wrong or missing CheXpert labels in predicted reports and asking the non-fine-tuned LLM for a corrected version.

\textbf{Natural Language Explanation:} Explain which part of the report indicates a specific pathology. We use the Mimic-NLE dataset~\cite{kayser2022explaining} as ground truth.

\textbf{View Classification:} Specify from which view (AP, PA, LL or lateral) the image was taken. The ground truth is collected from the metadata in the MIMIC-CXR dataset~\cite{johnson2019mimic}.

\subsection{Instruction Prompts}
We provide the exact prompts we use for report generation and three example prompts for the other tasks in the instruct dataset. The entire list of ten prompts per task will be included in our github repository.

\vspace{\baselineskip}
\noindent \textbf{Report Generation:}

\noindent Image information: \textless IMG\textgreater. Predicted Findings: \textless FINDINGS\textgreater. You are to act as a radiologist and write the finding section of a chest x-ray radiology report for this X-ray image and the given predicted findings. Write in the style of a radiologist, write one fluent text without enumeration, be concise and don't provide explanations or reasons.

\vspace{\baselineskip}
\noindent \textbf{Impression Generation:}

\noindent $\bullet$ What is the impression of this radiology report? \\
\noindent $\bullet$ Summarize the radiology report findings into an impression section. \\
\noindent $\bullet$ Can you formulate the impression section based on the radiology report's findings? 

\vspace{\baselineskip}
\noindent \textbf{Complete CheXpert QA:}

\noindent $\bullet$ List all the finding in this report. \\
\noindent $\bullet$ Enumerate the observations from the report. \\
\noindent $\bullet$ What findings can be identified from this report? 

\vspace{\baselineskip}
\noindent \textbf{Binary CheXpert QA:}

\noindent $\bullet$ Is there evidence of \textless PATHOLOGY\textgreater in the report? \\
\noindent $\bullet$ Is there any \textless PATHOLOGY\textgreater? \\
\noindent $\bullet$ Does the patient have \textless PATHOLOGY\textgreater? 

\vspace{\baselineskip}
\noindent \textbf{Rad-ReStruct QA:}

\noindent $\bullet$ [...] Answer with one of the following options: yes, no. Question: Is there pulmonary atelectasis in the lung? \\
\noindent $\bullet$ [...] From the given list, name all correct options: left lower lobe, left upper lobe, middle lobe, right lower lobe, right upper lobe. Question: In which part of the body? \\
\noindent $\bullet$ [...] From the given list, name all correct options: focal, no selection, patchy, round, scattered, small, streaky. Question: What are the attributes?

\vspace{\baselineskip}
\noindent \textbf{Region QA:}

\noindent $\bullet$ Is the patient's heart healthy? \\
\noindent $\bullet$ Does the patient have any abnormalities in the osseous structures? \\
\noindent $\bullet$ Are there any abnormalities in the lungs?

\vspace{\baselineskip}

\noindent \textbf{Easy Language:}

\noindent $\bullet$ Explain this report in very easy terms, such that a child would understand. \\
\noindent $\bullet$ Given this chest xray report, formulate it in easy language. \\
\noindent $\bullet$ Reformulate this report in simple and understandable language.

\vspace{\baselineskip}

\noindent \textbf{Summarization:}

\noindent $\bullet$ Summarize this report with bullet points. \\
\noindent $\bullet$ Provide a short summary of the most important points in this chest x-ray report. \\
\noindent $\bullet$ Please summarize this report in one sentence.

\vspace{\baselineskip}

\noindent \textbf{Correction:}

\noindent $\bullet$ The patient also has \textless PATHOLOGIES\textgreater, correct the report. \\
\noindent $\bullet$ There is no \textless PATHOLOGIES\textgreater, please adapt the report accordingly. \\
\noindent $\bullet$ I disagree with the generated report, I think the patient has \textless PATHOLOGIES\_1\textgreater, but does not have \textless PATHOLOGIES\_2\textgreater. Please adapt the report.

\vspace{\baselineskip}

\noindent \textbf{Natural Language Explanations:}

\noindent $\bullet$ What are the indicators for \textless PATHOLOGY\textgreater in the report?\\
\noindent $\bullet$ Why do you think the patient has \textless PATHOLOGY\textgreater? \\
\noindent $\bullet$ Which symptoms led to the diagnosis of \textless PATHOLOGY\textgreater?

\vspace{\baselineskip}

\noindent \textbf{View Classification:}

\noindent $\bullet$ From what view was this image taken? \\
\noindent $\bullet$ Can you identify the view of this x-ray image? \\
\noindent $\bullet$ Can you specify the angle from which this x-ray scan was taken?
\begin{figure}[tbh]
\centering
  \includegraphics[width=1.0\textwidth]{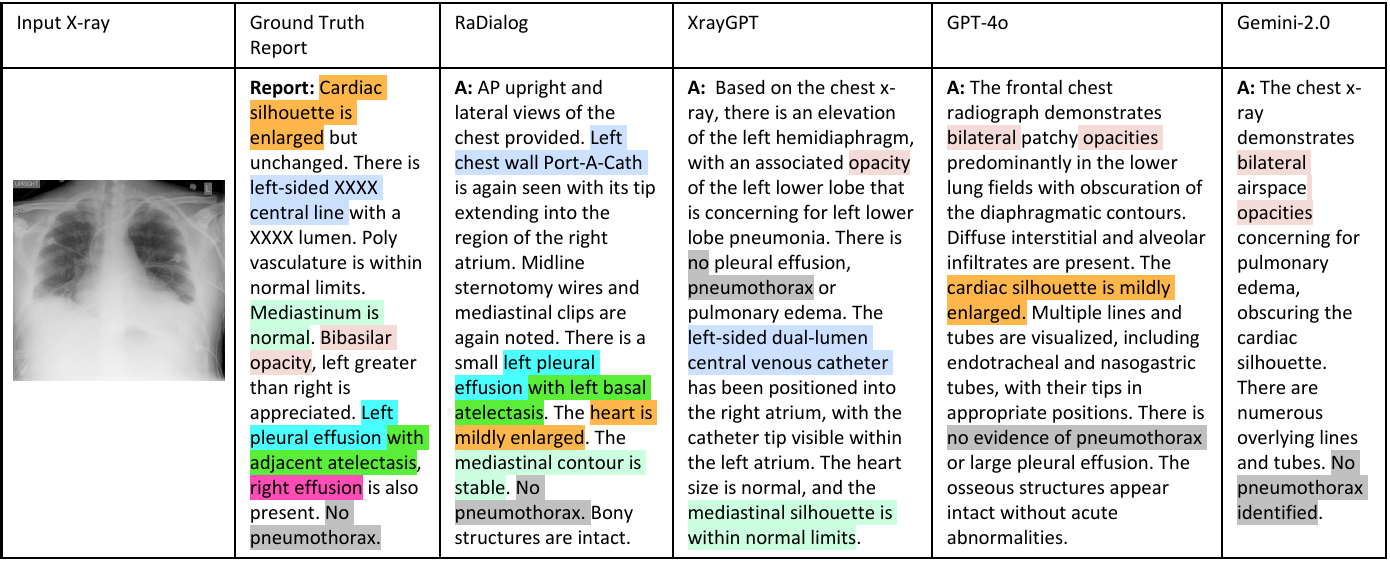}
   \caption{Qualitative report generation comparison of RaDialog with XrayGPT and GPT4-Vision.}
   \label{fig:chat_comp}
\end{figure}

\section{Additional Qualitative Results}
\label{ap:qual}
\begin{figure}[tb]
\centering
\includegraphics[width=0.8\linewidth]{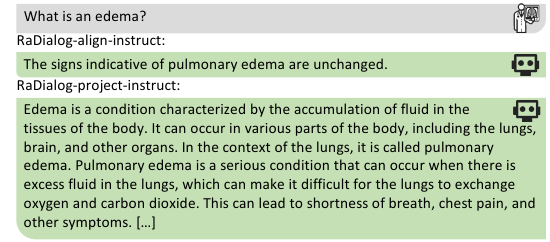}
\caption{Differences in conversation behavior of RaDialog-align-instruct and RaDialog-project-instruct in zero-shot conversational tasks.}
\label{fig:blipvsllava}
\end{figure}

In Fig. \ref{fig:chat_comp}, we provide a qualitative comparison of RaDialog's performance to XrayGPT \cite{thawkar2023xraygpt}, GPT-4o, and Gemini-2.0 on an out-of-domain image from the IU-Xray dataset \cite{iu-xray}, all prompted for report generation. The prompt details to write the findings section of a radiology report, in a concice style like a radiologist. All models are aware of the correct style for report writing, but XrayGPT gets fewer findings correct and hallucinates more, while both GPT-4o and Gemini-2.0 miss most of the findings, whereas RaDialog identifies almost all of them correctly. This underlines the importance of developing domain-specific models targeted at clinical correctness. Further, Fig.~\ref{fig:blipvsllava} shows an example of the benefits of RaDialog\textsubscript{project-ins} in interactive abilities in zero-shot knowledge question answering compared to RaDialog\textsubscript{align-ins}.

\section{Ablation studies}

\subsection{Indication-based Methods}
\label{ap:ind}
\begin{table}
  \centering 
  \caption{Comparison to MedPaLM~\cite{tu2024towards} and MAIRA-1~\cite{hyland2023maira}, both closed source models using indication (Ind.) as input, compared to training on publicly available data, allowing also to publish the model.}
  \label{tab:medpalm}
  \begin{tabular}{@{}lccccccc@{}}
    \hline
    Method & Public & Ind. & CE & B-1 & B-4 & R-L \\
    \hline
    MAIRA-1~\cite{hyland2023maira} & $\times$ & $\checkmark$ & 38.6 & \textbf{39.2} & 14.2 & 28.9\\
    MedPaLM-12b~\cite{tu2024towards} & $\times$ & $\checkmark$ & 37.3 & 30.9 & 10.4 & 26.2\\
    MedPaLM-84b~\cite{tu2024towards} & $\times$ & $\checkmark$ & \textbf{39.8} & 32.2 & 11.3 & 27.3\\
    RaDialog-align-report & $\checkmark$ & $\times$ & 39.4  & 34.6 & 9.5 & 27.1\\
    RaDialog-align-report & $\checkmark$ & $\checkmark$ & 39.2  & \textbf{39.2} & \textbf{14.8} & \textbf{31.6}\\
    \hline
  \end{tabular}
\end{table}
We compare our model to MedPaLM~\cite{tu2024towards} and MAIRA-1~\cite{hyland2023maira} in Tab.~\ref{tab:medpalm}. We separated this comparison because, unlike other state-of-the-art methods, these two use the indication section of the report as input. For comparison, we evaluate RaDialog with this additional input information and show that using the indication section leads to a significant jump in performance in the NLG metrics. Even though MedPaLM and MAIRA-1 rely on image and text encoders pre-trained with large-scale private data, we outperform MedPaLM-12b and MAIRA-1 (7b parameters) in all metrics and the 84b variant in the text-based metrics while having comparable clinical efficacy.

\subsection{Impact of Model Size}
\label{ap:size}
\begin{table}
  \centering
    \caption{Effect of different LLM sizes on report generation performance of RaDialog-align. Sec. denotes the average number of seconds to generate one report}

  \begin{tabular}{@{}lccccccc@{}}
    \hline
    LLM size & Sec. &CE & BS & B-1 & B-4 & MTR & R-L \\
    \hline
    Vicuna-7b & \textbf{1.2} & \textbf{39.4} & \textbf{0.40} & 34.6 & \textbf{9.5} & 14.0 & \textbf{27.1}\\
    Vicuna-13b & 1.9 & \textbf{39.4} & 0.39 & 34.8 & \textbf{9.5} & 14.0 & \textbf{27.1}\\
    Vicuna-33b & 7.9 & 39.0 & \textbf{0.40} & \textbf{35.0} & \textbf{9.5} & \textbf{14.1} & 27.0\\
    \hline
  \end{tabular}
  \label{tab:size}
\end{table}

Comparing different sizes of the LLM (Tab.~\ref{tab:size}), we observe that just scaling up the LLM size does not lead to a relevant performance increase, while leading to a slower inference time. Therefore, we opt to use the seven billion parameter version for our experiments, leading to faster training and inference speeds.

\section{Implementation Details}
\label{ap:imp}
We initialize our LLM with vicuna-7b~\cite{vicuna2023} and fine-tune it using LoRA~\cite{hu2021lora} with a learning rate (LR) of $3 \times 10^{-4}$ for one or four epochs (RaDialog-align-instruct/report) on a single Nvidia A-40 GPU with 48GB memory. For RaDialog-project, we use an LR of $2 \times 10^{-5}$ and train up to five epochs with early stopping. BioVil-T~\cite{bannur2023learning} is fine-tuned for multi-label classification of CheXbert findings~\cite{smit2020chexbert} using log-weighted cross-entropy loss (six epochs, LR $5 \times 10^{-5}$) and employed for visual feature extraction. RaDialog-align uses BERT~\cite{devlin2018bert} to align text and image features, trained with cosine annealing LR ($1 \times 10^{-5}$ to $1 \times 10^{-4}$) and linear warmup over four epochs.

\end{document}